\documentclass[conference]{IEEEtran}
\IEEEoverridecommandlockouts
\usepackage{cite}
\usepackage{amsmath,amssymb,amsfonts}
\usepackage{algorithmic}
\usepackage{graphicx}
\usepackage{textcomp}
\usepackage{xcolor}
\usepackage{lipsum}
\usepackage{multicol}
\usepackage{subcaption}
\usepackage{soul}
\usepackage{bm}
\usepackage{float}
\usepackage{subcaption}
\usepackage{tabularx}
\newcolumntype{Y}{>{\centering\arraybackslash}X}
\RequirePackage{enumitem}

\setlist[enumerate]{itemsep=0pt, topsep=0pt, itemindent=15pt, leftmargin=0pt,listparindent=\parindent}
\setlist[itemize]{itemsep=0pt, topsep=0pt, itemindent=15pt, leftmargin=0pt,listparindent=\parindent}

\newcommand{\ignore}[1]{}

\def\BibTeX{{\rm B\kern-.05em{\sc i\kern-.025em b}\kern-.08em
    T\kern-.1667em\lower.7ex\hbox{E}\kern-.125emX}}
\begin{document}
\title{Bayesian Inference Accelerator for Spiking Neural Networks}
 {
\author{\IEEEauthorblockN{Prabodh Katti$^1$, Anagha Nimbekar$^{1,2}$, Chen Li$^1$, Amit Acharyya$^2$, Bashir M. Al-Hashimi$^1$, Bipin Rajendran$^1$}
\IEEEauthorblockA{
$^1$King's College London, London, UK; $^2$IIT Hyderabad, Hyderabad, India \\
Email: bipin.rajendran@kcl.ac.uk}
}}

\maketitle

\begin{abstract}
Bayesian neural networks offer better estimates of model uncertainty compared to frequentist networks. However,  inference involving Bayesian models requires multiple instantiations or sampling of the network parameters, requiring significant computational resources. Compared to traditional deep learning networks, spiking neural networks (SNNs) have the potential to reduce computational area and power, thanks to their event-driven and spike-based computational framework. Most works in literature either address frequentist SNN models or non-spiking Bayesian neural networks. In this work, we demonstrate an optimization framework for developing and implementing efficient Bayesian SNNs in hardware by additionally restricting network weights to be binary-valued to further decrease power and area consumption.  We demonstrate accuracies comparable to Bayesian binary networks with full-precision Bernoulli parameters, while requiring up to $25\times$ less spikes than equivalent binary SNN implementations. We show the feasibility of the design by mapping it onto Zynq-7000, a lightweight SoC, and achieve a $6.5 \times$ improvement in GOPS/DSP while utilizing up to 30 times less power compared to the state-of-the-art. 

\end{abstract}


\section{Introduction}
Modern neural networks tend to produce overconfident and poorly calibrated results. This is especially true of large and deep models \cite{guo2017calibration}. With deep learning systems getting deployed for critical and sensitive applications, assigning uncertainty measures appropriately makes the overall decision-making more trustworthy  \cite{Kendall_Gal_2017}. The hallmark of a well-calibrated classification model is that the accuracy of a decision should match the probability estimate for the predicted class \cite{guo2017calibration}.

\begin{figure}[!h]
\centering
 \includegraphics[width=.35\textwidth]{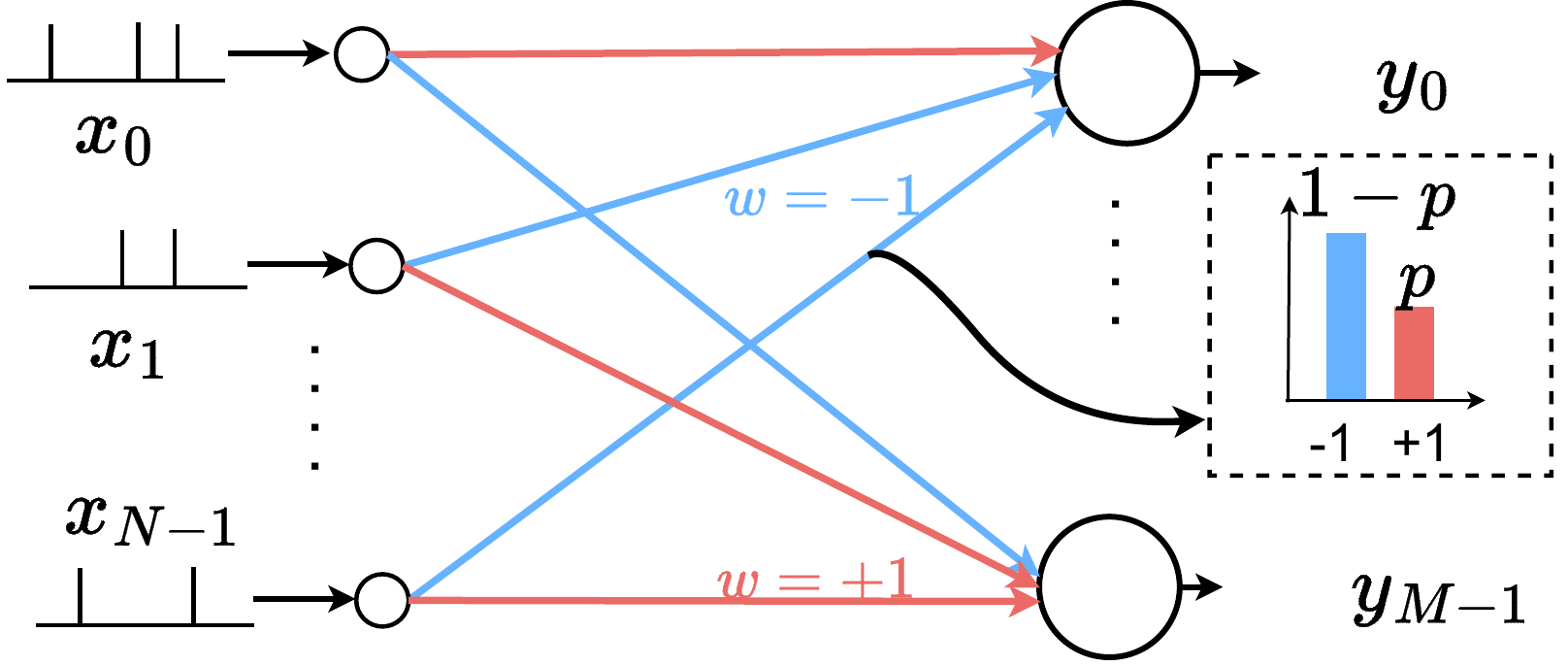}

   \centering \includegraphics[width=.45\textwidth]{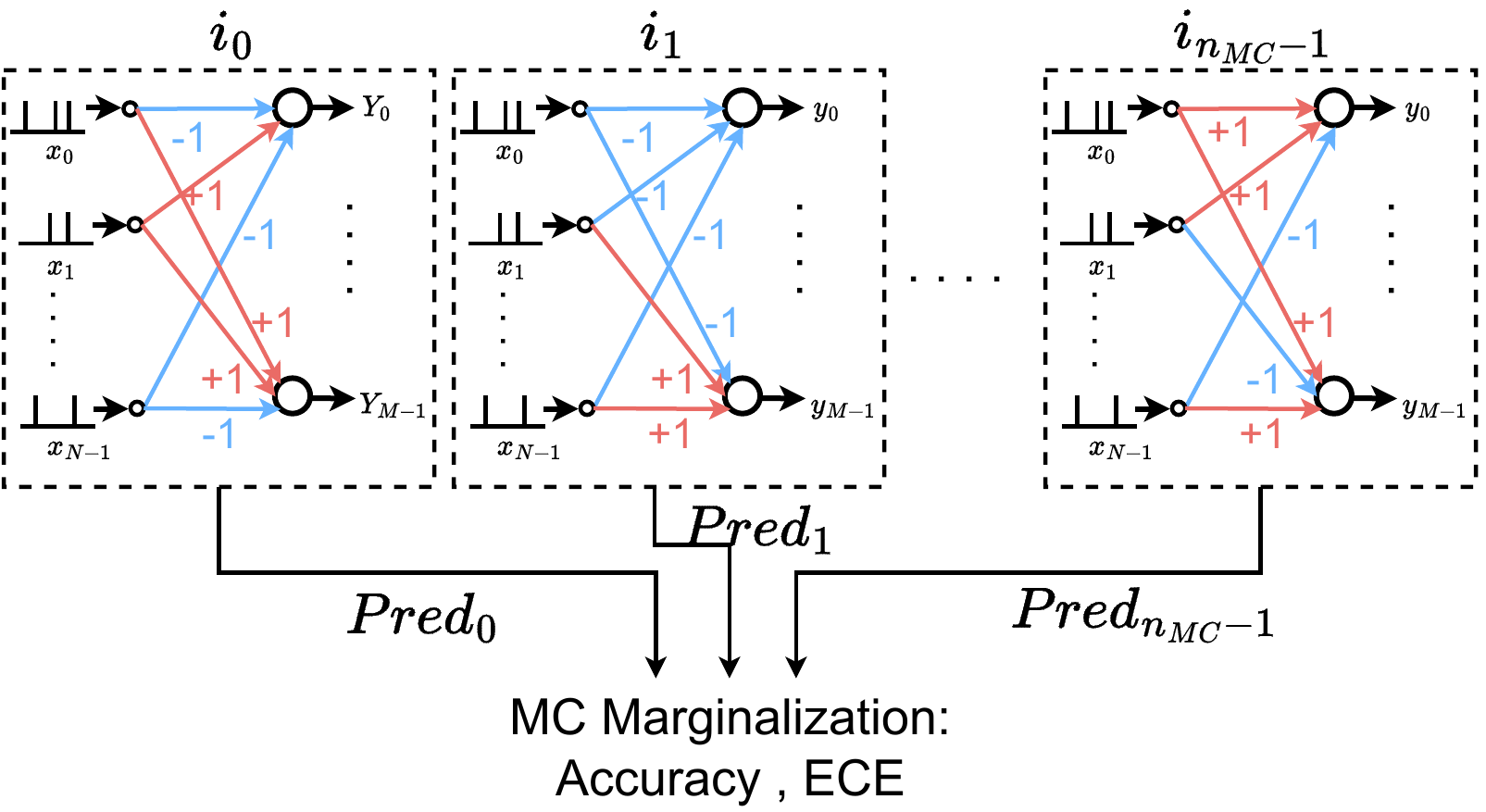}

\caption{Top: Bayesian SNN with Bernoulli distributed weights  (i.e., weights take binary values upon sampling). Bottom: Bayesian inference is performed by the Monte Carlo (MC) method, where the network is sampled $n_{MC}$ times and the predictions are combined to get the final result.  } 
\label{fig:sampling}
\end{figure}

Bayesian neural networks have been shown to produce better-calibrated predictions compared to their frequentist counterparts by encoding model uncertainty in probability distributions of the network parameters \cite{jang2021bisnn}. Multiple ensembles of a Bayesian network can be used to generate a set of predictions that are marginalized to get a better-calibrated decision (Fig.~\ref{fig:sampling}). This however makes their deployment challenging on power and size-constrained edge systems. We consider the problem of achieving a hardware-friendly Bayesian accelerator that matches the classification performance of a state-of-the-art software implementation. Towards this goal, we make the following design choices in this study: 
\begin{itemize}
    \item Create ensembles in time rather than in space, i.e., generating multiple samples of the physical network by `instantiating' it sequentially, as opposed to creating multiple physical samples of the network in hardware.
    \item Use spiking neural networks (SNNs), as they encode inputs as well as post-neuron outputs as binary all-or-none signals (`spikes'), instead of using real numbers as is done in modern deep artificial neural network (ANN) models. 
    \item Use network weights that are Bernoulli distributed, i.e., the trained Bernoulli parameter for each weight generates only two states $\{ -1, +1\}$ upon sampling during inference, to further reduce hardware implementation complexity.
    \item Develop a custom ANN-to-SNN training and conversion method that is  applicable to state-of-the-art deep Bayesian networks optimized for reduced-precision and low-latency  (i.e., number of spikes needed) operation. 
  \end{itemize}

\vspace{-0.01in}
\noindent \emph{Related Work:} ANN-to-SNN conversion methods for low latency inference have been proposed in \cite{bu2023optimal} and \cite{10.3389/fnins.2022.918793}, with  software-equivalent accuracy demonstrated for  ResNet-18 on CIFAR-10 dataset within 8 steps. These methods however require full-precision weights and have not been applied for Bayesian networks. 
The special case of a frequentist binary SNN was explored in \cite{lu2020exploring},  but over 50 timesteps were necessary to achieve software-equivalent accuracy with VGG-15 network on CIFAR-100 dataset. Others either use non-uniform timesteps in layers \cite{dinh2023nuts}, or have a few full-precision \cite{pei2023albsnn} layers, leading to increased implementation complexity.

A  naive Bayesian classifier adapted for SNNs was demonstrated in  \cite{9585371}, but it uses a hierarchical SNN model and not a Bayesian neural network.
Most hardware implementations of  Bayesian neural networks   focus on non-spiking architectures, and have considered methods such as  MC dropout \cite{10.1109/DAC18074.2021.9586137}, use small datasets with MLP-only implementation \cite{10.1145/3173162.3173212} or use resource-intensive Gaussian random numbers generated from Bernoulli generators  \cite{AWANO20231}. We develop a novel hardware-software co-design framework to design accelerators for deep Bernoulli Bayesian SNNs that can provide fast and accurate decisions with quantifiable metrics of calibration.

\section{Background}\label{sec: Background}




\noindent \emph{Bayesian Bernoulli neural network:}
The weights $\boldsymbol{w}$ of the Bayesian neural network considered here are Bernoulli distributed, i.e each element in the weight vector $\boldsymbol{w}$ only takes the values $\{+1,-1\}$, with $p_w=Pr(w=1)$ being the Bernoulli parameter (Fig. \ref{fig:sampling}). Following the approach described   in \cite{Meng_Bachmann_Khan_2020}, during training, the natural parameters of the model $\boldsymbol{\lambda}$ (a vector that is a collection of individual natural parameter $\lambda_w$) are obtained from training by variational inference using BayesBiNN optimizer. $\lambda_w$ is associated with the Bernoulli parameter $p_w$ according to
\begin{equation}
    p_w = \mathrm{\sigma}(2\lambda_w),
    \label{eq:bern_parameter}
\end{equation}
where $\sigma(\cdot)$ is the sigmoid function.

\begin{figure}
  \centering
    \includegraphics[width=0.3\textwidth]{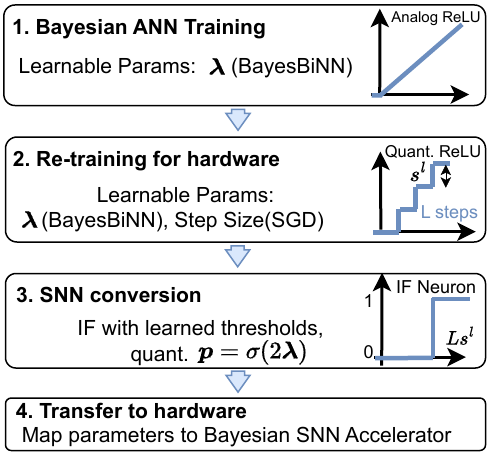}
    \caption{Our optimization methodology to develop   Bayesian SNNs for hardware implementation. }
  \label{fig:hwswopt}
\end{figure}

Once the network is trained, we utilize the learnt $\boldsymbol{p}$ to generate network ensembles. The spike outputs of the networks are accumulated and processed (softmax) to obtain a well-calibrated result, as shown in Fig. \ref{fig:sampling}.

\noindent \emph{Spiking neural networks:}
In SNNs, inputs are presented in the form of spike trains $\boldsymbol{z_{j,t}}$ lasting  $T$ timesteps, where $j$ denotes the index of the neuron that emits the spike and $t$ denotes the timestep. The emitted spike propagates through the synaptic weight $w_{ij}$ before arriving at neuron $i$. Each neuron $i$ maintains an internal state called membrane potential $U_i$ and emits its spike according to the condition
\begin{equation}
    z_{i,t+1} = \Theta(U_{i,t} - \theta_i),
    \label{eq:heaviside}
\end{equation}
where $\Theta(\cdot)$ is the Heaviside  function and $\theta_i$ is the threshold of neuron $i$. Upon emission of a spike, the membrane potential resets either to zero or soft resets to a smaller value, referred to as reset-by-subtraction \cite{bu2023optimal}. We use reset-by-subtraction and integrate-and-fire (IF) neurons in our work. Spikes are encoded from real-world data via rate-coding, with the encoder being a trained convolutional layer.
\section{Methodology}
\begin{figure}
\centering
        \includegraphics[width=0.4\textwidth]{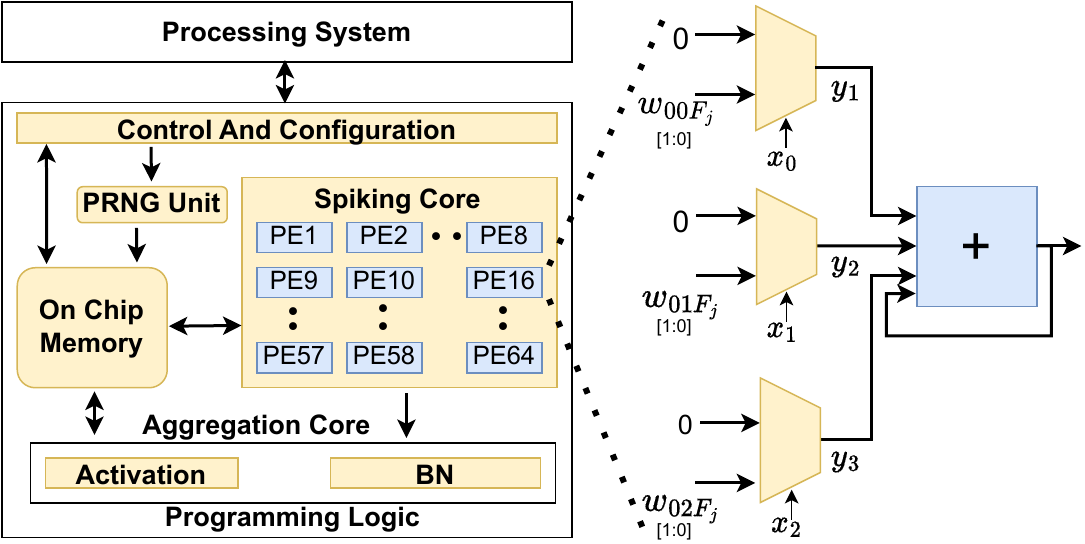}
    
    \caption{Design of the novel inference architecture for Bayesian binary SNNs.}
    \label{fig:Arch}
\end{figure}

We propose a framework (Fig.~\ref{fig:hwswopt}) to implement Bayesian SNNs in a novel inference accelerator architecture (Fig.~\ref{fig:Arch}) comprising of the following steps:
(1) Train a Bayesian binary ANN in software with full-precision Bernoulli parameters \cite{Meng_Bachmann_Khan_2020}. 
(2) Re-train the ANN for reduced precision hardware implementation by replacing the software ReLU by a quantized version. Retraining is necessary to ensure that there is no significant  drop in accuracy.
(3) Convert the trained ANN to SNN with reduced precision parameters.
(4) Map the optimized SNN into the hardware and experimentally validate the approach by quantifying classification accuracy and uncertainty as well as hardware performance metrics. 

\subsection{Software optimization of Bayesian SNNs}
We introduce an ANN-to-SNN conversion method for Bayesian neural networks, by adopting methods described in \cite{bu2023optimal, Meng_Bachmann_Khan_2020}. First, the ANN is trained with ReLU activation, followed by an optimization step with a quantized ReLU with  $L$ levels, each with a trainable step size $s^l$ for a given layer $l$.  Post optimization, we obtain the equivalent SNN by replacing the quantized ReLU activation with IF activation. Networks with smaller values of $L$ require a smaller number of timesteps to reach the maximum possible accuracy, however, this comes at the cost of reduced model capacity as reflected in the overall classification accuracy. 
\ignore{
\begin{equation}
    y^l = s^l \times \bigg\lfloor \frac{1}{L} \frac{v^{l}L}{s^l}, 0, 1 \bigg\rceil
    \label{ANN-SNN}
\end{equation}
Where $v^{l}$ is the pre-activation given by $BN(W^lz^{l-1} + B^l)$, with $W^l$, $B^l$ and $z^{l-1}$ being weight, bias, and input spike to the layer $l$ respectively and $BN$ is the Batch Normalization.}

\subsection{Inference accelerator design for Bayesian SNN}\label{sec:architecture}
The details of the proposed Bayesian Bernoulli SNN hardware architecture are as follows:

\noindent \emph{Top level architecture}: 
In CNNs, convolution is the most critical operation, e.g., comprising about 90\% of all the operations for   Alex-Net \cite{Suda_Chandra_Dasika_Mohanty_Ma_Vrudhula_Seo_Cao_2016}, and more than 99.99\% for   ResNet-18   \cite{lu2020exploring}. These operations involve multiplication and accumulation (MAC), which are both area and power-intensive. Therefore, custom accelerators have been proposed utilizing a Processing System (PS) -- Programming Logic (PL) paradigm \cite{9908073,Suda_Chandra_Dasika_Mohanty_Ma_Vrudhula_Seo_Cao_2016,8954866}, where the PS performs the sequential functions such as image sliding and controlling the data flow, and the PL executes parallelizable functions such as weight sampling, convolution, and batch-normalization. We focus our attention on the custom PL unit implementing the Bayesian spiking computations and utilize a standard ARM Cortex-A9 processor for PS (Fig.~\ref{fig:Arch}).  
  


\noindent \emph{PRNG reuse architecture}:
To sample the network, we must generate Bernoulli distributed weights $\boldsymbol{w}$ corresponding to their Bernoulli parameter $\boldsymbol{p}$ as per    \eqref{eq:bern_parameter}. Towards this, we use an 8-bit quantized version of $\boldsymbol{p}$ obtained from software training and compare it with an 8-bit random number (RN) $\boldsymbol{r} \sim \mathcal{U}(0,1)$\cite{Sinharay_2010} generated in hardware from a custom pseudo-random number generator (PRNG) designed using linear feedback shift registers (LFSR) \cite{saluja1987linear}. \ignore{Each sampled binary weights is then generated as follows:
\begin{equation}
    w = 
     \begin{cases}
        +1 &\quad \text{if }r < p,\\
       -1 &\quad \text{if } r > p\\
     \end{cases}
    \label{eq:binarization}
\end{equation}}


\begin{figure}
\centering
     
    \includegraphics[width=.49\linewidth]{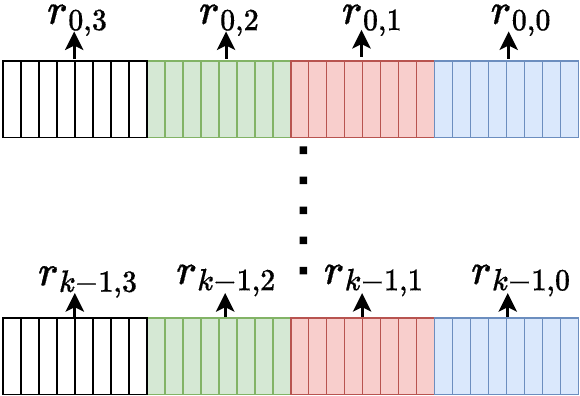}
  \hfill
    \includegraphics[width=.29\linewidth]{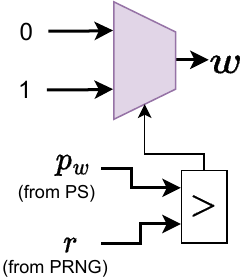}
 

   \caption{(Left) Resource-efficient design of an LFSR-based  PRNG with maximal reuse. Shown here are  $k$ rows and four 8-bit pseudo-random numbers taken from each 32-bit LFSR. As our design needs 64 RNs in a single clock, $k=16$ such LFSRs are sufficient. RNs generated from the PRNG block are then utilized by the Bernoulli RN generating block (right) to generate the Bernoulli weight.  A  two-bit representation is used for the product of \{-1,+1\} weights with \{0,1\} spikes.}
   \label{fig:prng}
\end{figure}

 Traditionally, the lowest 8 bits are taken from an LFSR to get a random number \cite{AnalogDevices}. To improve area efficiency, we tap all 4 bytes of an LFSR, thus yielding 4 random numbers from a single 32-bit LFSR. To sample all 64 weights transferred from PS into weight memory in a single clock cycle, we create a PRNG bank as shown in Fig.~\ref{fig:prng} with 16 LFSRs. This may result in some correlation between two adjacent random number bytes, but we empirically observe this to have minimal impact on the results, as discussed in Section \ref{sec:results} (Table \ref{table:rand}).

\noindent \emph{Processing elements (PE)}: The fundamental unit of convolution processing is the processing element, which performs the MAC operation. Because we are only utilizing spiking inputs, we can replace multipliers with multiplexers. We use the input spike as the `select' signal to choose between the weight $w$ and $0$, effectively implementing input-weight multiplication. Additionally, since $w \in \{-1,+1\}$, we can represent these with just 2 bits, further improving our resource efficiency. PE of a filter $F_j$ processes three rows of a filter in a single clock cycle. For the most common filter dimension of $3 \times 3$,  all the MAC operations are completed in 3 clock cycles. We utilize  64 PEs for parallel processing of 64 filters. Once complete, the results are stored in temporary buffer memories before being taken up for neuron processing.

\noindent \emph{Neurons}: Neurons perform batch normalization (BN) and integrate-and-fire (IF) operation. We optimize the BN operation by transforming the coefficients such that only two parameters are required to specify the entire operation, to further reduce the number of multiplication and addition operations.
Since we are reusing the four neurons among PE elements, we use 64 kB memory to store neuron states.


\section{Experiments and results} \label{sec:results}
We demonstrate the overall approach proposed in the paper by developing a deep Bayesian ResNet-18 ANN, converting it to 8-bit reduced-precision SNN, and finally implementing and quantifying the performance of this network architecture on an SoC. 

\begin{figure}[!h]
  \centering
    \includegraphics[width=0.39\textwidth]{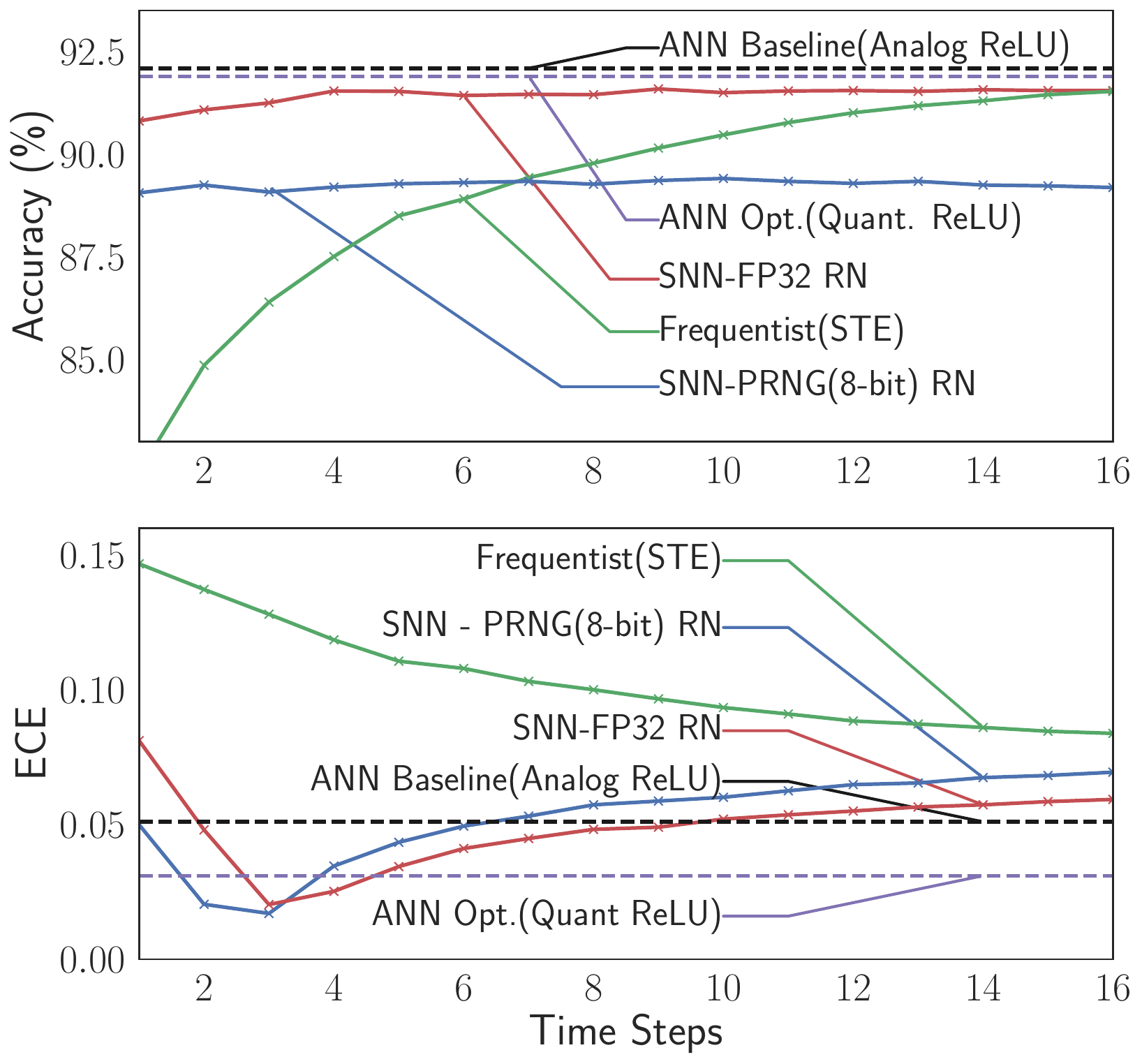}
    \caption{Classification performance (top) and ECE (bottom) on the CIFAR-10 dataset at various stages of optimization discussed in Fig.~\ref{fig:hwswopt} as well as of a frequentist counterpart.} 
  \label{snnresult}
\end{figure}

\begin{figure}
\centering    \includegraphics[width=0.4\textwidth]{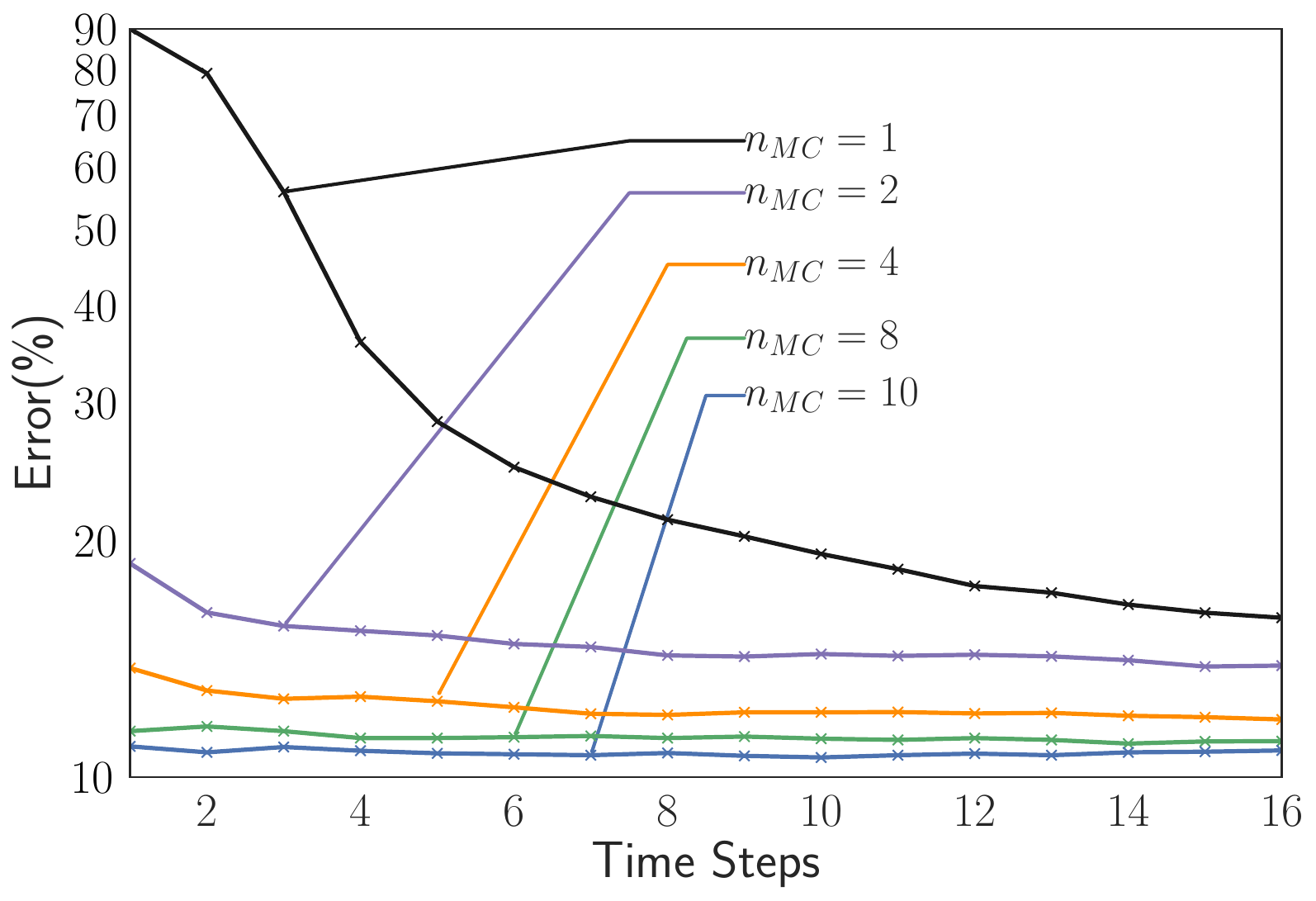}
    \caption{Inference error as a function of  SNN timesteps for the Bayesian SNN with PRNG RNs for various Monte-Carlo sampling instances $n_{MC}$.}
    \label{fig:nMC}
  \end{figure}
  
Our software results (Figs.~\ref{snnresult}, \ref{fig:nMC}) are based on the Bayesian training of a ResNet-18 network on the CIFAR-10 dataset.
ResNet-18 is an 18-layer deep convolutional network with residual connections with more than 11 Million parameters \cite{he2015deep}. Training Binary ANNs are challenging as they are slower to converge \cite{Alizadeh2018AnES}, as are Bayesian neural networks using variational inference \cite{le2022adaste} which additionally have noisier gradient descent \cite{9756596}. For ANN training, we utilized the hyperparameters available from \cite{Meng_Bachmann_Khan_2020} and \cite{le2022adaste}, and achieved an accuracy of $92.4\%$, close to the one achieved by \cite{le2022adaste} after nearly 8000 epochs of training. During the intermediate step of ANN-to-SNN conversion with BayesBiNN optimizer and SGD for the activation step size training, choosing the quantization level appropriately is crucial -- we observe significant accuracy loss using $L=2$ or $L=4$ as suggested in \cite{bu2023optimal} and \cite{10.3389/fnins.2022.918793}, but with $L=8$, we achieved an accuracy of $91.9\%$ in about 2000 epochs. We then built an 8-bit inference model by quantizing $\boldsymbol{p}$, batchnorm coefficients, and intermediate results, and evaluated inference accuracy with $n_{MC}=10$. We observe that within 4 timesteps, the SNN classification accuracy stabilizes to around $91.5\%$ (Fig. \ref{snnresult}), which is at least $12.5-25\times$ fewer timesteps than \cite{lu2020exploring}. The overall number of timesteps for an area-efficient design point considering ensembling in time will scale linearly with $n_{MC}$.

We also compare the performance of Bayesian networks with a straight-through-estimator-based (STE) frequentist binary network \cite{Meng_Bachmann_Khan_2020,jang2021bisnn,le2022adaste} in Fig. \ref{snnresult}. This network was trained using a standard ADAM optimizer and converted to SNN as described earlier. We note that the Bayesian SNN with FP32 random numbers not only matches STE accuracy but reaches maximum accuracy in fewer timesteps. 
We also observe that the expected calibration error (ECE), a measure of discrepancy between accuracy and confidence \cite{guo2017calibration}, is lower for the Bayesian SNN than its frequentist counterpart. In fact, even though the PRNG RN-based Bayesian SNN has poorer accuracy performance compared to STE, it still has better (lower) ECE. We notice a slow increase of ECE for Bayesian SNNs after 4 timesteps. This happens because while the confidence, i.e., $Pr(\mathrm{correct\,classification})$ keeps accumulating with timesteps, its  effect on accuracy increase is minimal \cite{li2023unleashing}. Hence, we can  stop computations early to get a well-calibrated prediction, thus making a compelling case for the use of Bayesian SNNs for trustworthy decision-making  (See Fig. \ref{fig:nMC}).

During the ANN-to-SNN conversion, we observed that accuracy drops to around $89.4\%$ upon using the hardware-friendly 8-bit PRNG. To understand the tradeoffs better, we compare the network performance between four strategies for RN generation: using $rand(\cdot)$ and $randint(\cdot)$ function from software libraries, 8-bit LFSR with no reuse i.e., only 1 RN used from each LFSR and the LFSR with maximal reuse as discussed earlier. We observe that reusing LFSR bits has minimal impact on network accuracy, but bit precision of the random number has a significant impact, as evidenced by the drop in accuracy while using the $randint(\cdot)$ function. Random numbers of higher-bit precision sample the sample-space more effectively, but this comes at the cost of hardware resources and execution time (Table \ref{table:rand}).
  \begin{table}[!h]
  \caption{Network accuracy and hardware execution time  (measured on the PYNQ board for generating 64 RNs) for the different schemes for Bernoulli random number generation.}
    \label{table:rand}
    \centering
      \begin{tabularx}{\columnwidth}{|c|Y|Y|Y|Y|}
      \hline
     \emph{\textbf{Implementation}}  & \emph{\textbf{Clk Freq}}&\emph{\textbf{Time  (ms)}} & \emph{\textbf{Accuracy (\% @T=16)}} \\
     \hline
     \textbf{FP32}, SW  & - & - & $91.5$ \\
     \hline
     \textbf{FxP8 (randint())}, SW  & $665$ MHz & $1.5$ & $89.32$ \\
     \hline
     \textbf{LFSR (no reuse)}, HW  & $94$ MHz & $1.06\times10^{-5}$ & $89.64$ \\
     \hline
     \textbf{LFSR (max reuse)},HW &   $94$ MHz & $1.06\times10^{-5}$ &$89.41$ \\
      \hline
    \end{tabularx}

  \end{table}

\begin{table}
\begin{center}
\caption{FPGA resource utilization.}
\label{tab:util}
\begin{tabular}{|c|c|c|c|}
\hline
\textbf{Parameter} & \textbf{Utilized} & \textbf{Available} & \textbf{Percentage} \\
\hline
LUTs & 2819 & 53200  & 5.3\%  \\
\hline
FFs & 2962 & 106400  & 2.78\%  \\
\hline
DSPs & 6 & 220 & 2.73\% \\
\hline
BRAMs & 47 & 140 & 33.57\% \\  \hline
LUTRAMs & 60 & 17400 & 0.34\%  \\ \hline 
BUFG & 1 & 32 & 3.13\%  \\\hline 

\end{tabular}

\end{center}
\end{table}

\begin{table}[!h]
\centering
\caption{Comparison with other works in literature.}
\label{table:HW}
\begin{tabularx}{\columnwidth}{|c|Y|Y|Y|Y|} \hline
{\textbf{Work}}&{\textbf{\cite{Jia2020EfficientCR}}} & {\textbf{\cite{10.1145/3173162.3173212}}} & {\textbf{\cite{10.1109/DAC18074.2021.9586137}}} & \textbf{This work} \\ \hline 
Implementation & ASIC & FPGA & FPGA & FPGA \\ \hline 
Network & MLP & MLP & ResNet-18 & ResNet-18 \\ \hline 
Spiking & N & N & N & Y \\ \hline 
Process & 45nm & - & - & -  \\ \hline 
Area (mm$^2$) & 6.63 & - & - & -  \\ \hline 
Freq. (MHz) & - & 212.9 & 225 & 94  \\ \hline 
Wt Precision & 8 & 8 & 8 & 8 \\ \hline 
Power (W) & 0.5 & 6.11 & 45 & 1.47\\ \hline 
Throughput (GOPS) & 1.86 & 59.57 & 1590 & 42.1  \\ \hline 
Energy Eff. (GOPS/W)& 3.72 & 9.75 & 35.33& 28.64 \\ \hline
Area Eff. (GOPS/mm$^2$) & 0.561 & - & - & - \\ \hline
GOPS/DSPs & - & 0.174 & 1.079 & 7.01 \\  \hline

\end{tabularx}

\end{table}



We implement the proposed architecture on a PYNQ SoC platform to evaluate hardware performance (See Table \ref{tab:util}). PYNQ's ARM Cortex processor is used as PS. To the best of our knowledge, Bayesian binary SNN implementations have not been reported so far; hence, we compare  our accelerator with other  Bayesian implementations and compare peak throughput (Giga Operations/s or GOPS), energy, and area efficiency (Table \ref{table:HW}). The achieved energy efficiency is comparable to \cite{10.1109/DAC18074.2021.9586137}, and outperforms other FPGA implementations in hardware resources (GOPS/DSP slices), thanks to the optimized multiplier-less PEs in our implementation. We report the total  power consumed by both PS and PL.

\section{Conclusions}

This paper demonstrates a  Bayesian spiking neural network optimized for efficient hardware implementation that provides high accuracy and appropriately confident predictions within 4 timesteps, providing a novel solution that addresses the overconfidence problem of modern deep neural networks. We also designed and demonstrated a  hardware-friendly reduced-precision Bayesian SNN model on a custom accelerator platform leveraging optimized reuse of random number generators, identifying performance trade-offs. The overall system throughput is limited by the von Neumann bottleneck of the processor-programming logic data transfer, which can be addressed through in-memory computing architectures based on  CMOS or emerging memory technologies.

\section*{Acknowledgment}

  This research was supported in part by the EPSRC Open Fellowship EP/X011356/1 and the EPSRC grant EP/X011852/1. We also thank MeitY, Government of India, for their support.
\bibliography{Bayesian_Inference_for_SNN}

\begin{thebibliography}{10}
\providecommand{\url}[1]{#1}
\csname url@samestyle\endcsname
\providecommand{\newblock}{\relax}
\providecommand{\bibinfo}[2]{#2}
\providecommand{\BIBentrySTDinterwordspacing}{\spaceskip=0pt\relax}
\providecommand{\BIBentryALTinterwordstretchfactor}{4}
\providecommand{\BIBentryALTinterwordspacing}{\spaceskip=\fontdimen2\font plus
\BIBentryALTinterwordstretchfactor\fontdimen3\font minus \fontdimen4\font\relax}
\providecommand{\BIBforeignlanguage}[2]{{%
\expandafter\ifx\csname l@#1\endcsname\relax
\typeout{** WARNING: IEEEtran.bst: No hyphenation pattern has been}%
\typeout{** loaded for the language `#1'. Using the pattern for}%
\typeout{** the default language instead.}%
\else
\language=\csname l@#1\endcsname
\fi
#2}}
\providecommand{\BIBdecl}{\relax}
\BIBdecl

\bibitem{guo2017calibration}
C.~Guo, G.~Pleiss, Y.~Sun, and K.~Q. Weinberger, ``On calibration of modern neural networks,'' in \emph{International conference on machine learning}.\hskip 1em plus 0.5em minus 0.4em\relax PMLR, 2017, pp. 1321--1330.

\bibitem{Kendall_Gal_2017}
A.~Kendall and Y.~Gal, ``What uncertainties do we need in {Bayesian} deep learning for computer vision?'' in \emph{Proceedings of the 31st International Conference on Neural Information Processing Systems}, ser. NIPS'17.\hskip 1em plus 0.5em minus 0.4em\relax Red Hook, NY, USA: Curran Associates Inc., 2017, p. 5580–5590.

\bibitem{jang2021bisnn}
H.~Jang, N.~Skatchkovsky, and O.~Simeone, ``{{BiSNN}: Training spiking neural networks with binary weights via Bayesian learning},'' in \emph{2021 IEEE Data Science and Learning Workshop (DSLW)}.\hskip 1em plus 0.5em minus 0.4em\relax IEEE, 2021, pp. 1--6.

\bibitem{bu2023optimal}
\BIBentryALTinterwordspacing
T.~Bu, W.~Fang, J.~Ding, P.~DAI, Z.~Yu, and T.~Huang, ``Optimal {ANN}-{SNN} conversion for high-accuracy and ultra-low-latency spiking neural networks,'' in \emph{International Conference on Learning Representations}, 2022. [Online]. Available: \url{{https://openreview.net/forum?id=7B3IJMM1k\_M}}
\BIBentrySTDinterwordspacing

\bibitem{10.3389/fnins.2022.918793}
\BIBentryALTinterwordspacing
C.~Li, L.~Ma, and S.~Furber, ``{Quantization Framework for Fast Spiking Neural Networks},'' \emph{Frontiers in Neuroscience}, vol.~16, 2022. [Online]. Available: \url{https://www.frontiersin.org/articles/10.3389/fnins.2022.918793}
\BIBentrySTDinterwordspacing

\bibitem{lu2020exploring}
S.~Lu and A.~Sengupta, ``Exploring the connection between binary and spiking neural networks,'' \emph{Frontiers in neuroscience}, vol.~14, p. 535, 2020.

\bibitem{dinh2023nuts}
V.-N. Dinh, N.-M. Bui, V.-T. Nguyen, D.~John, L.-Y. Lin, and Q.-K. Trinh, ``{{NUTS-BSNN}: A Non-uniform Time-step Binarized Spiking Neural Networks with Energy-Efficient In-memory Computing Macro},'' \emph{Neurocomputing}, p. 126838, 2023.

\bibitem{pei2023albsnn}
Y.~Pei, C.~Xu, Z.~Wu, Y.~Liu, and Y.~Yang, ``{{ALBSNN}: ultra-low latency adaptive local binary spiking neural network with accuracy loss estimator},'' \emph{Frontiers in Neuroscience}, vol.~17, 2023.

\bibitem{9585371}
B.~Deng, Y.~Fan, J.~Wang, and S.~Yang, ``{Reconstruction of a Fully Paralleled Auditory Spiking Neural Network and {FPGA} Implementation},'' \emph{IEEE Transactions on Biomedical Circuits and Systems}, vol.~15, no.~6, pp. 1320--1331, 2021.

\bibitem{10.1109/DAC18074.2021.9586137}
\BIBentryALTinterwordspacing
H.~Fan, M.~Ferianc, M.~Rodrigues, H.~Zhou, X.~Niu, and W.~Luk, ``{High-Performance FPGA-Based Accelerator for Bayesian Neural Networks}.''\hskip 1em plus 0.5em minus 0.4em\relax IEEE Press, 2021, p. 1063–1068. [Online]. Available: \url{https://doi.org/10.1109/DAC18074.2021.9586137}
\BIBentrySTDinterwordspacing

\bibitem{10.1145/3173162.3173212}
\BIBentryALTinterwordspacing
R.~Cai, A.~Ren, N.~Liu, C.~Ding, L.~Wang, X.~Qian, M.~Pedram, and Y.~Wang, ``{VIBNN: Hardware Acceleration of Bayesian Neural Networks},'' ser. ASPLOS '18.\hskip 1em plus 0.5em minus 0.4em\relax New York, NY, USA: Association for Computing Machinery, 2018, p. 476–488. [Online]. Available: \url{https://doi.org/10.1145/3173162.3173212}
\BIBentrySTDinterwordspacing

\bibitem{AWANO20231}
\BIBentryALTinterwordspacing
H.~Awano and M.~Hashimoto, ``{B2N2: Resource efficient Bayesian neural network accelerator using Bernoulli sampler on FPGA},'' \emph{Integration}, vol.~89, pp. 1--8, 2023. [Online]. Available: \url{https://www.sciencedirect.com/science/article/pii/S0167926022001523}
\BIBentrySTDinterwordspacing

\bibitem{Meng_Bachmann_Khan_2020}
X.~Meng, R.~Bachmann, and M.~E. Khan, ``Training binary neural networks using the bayesian learning rule,'' in \emph{International conference on machine learning}.\hskip 1em plus 0.5em minus 0.4em\relax PMLR, 2020, pp. 6852--6861.

\bibitem{Suda_Chandra_Dasika_Mohanty_Ma_Vrudhula_Seo_Cao_2016}
\BIBentryALTinterwordspacing
N.~Suda, V.~Chandra, G.~Dasika, A.~Mohanty, Y.~Ma, S.~Vrudhula, J.-s. Seo, and Y.~Cao, ``{Throughput-Optimized OpenCL-based FPGA Accelerator for Large-Scale Convolutional Neural Networks},'' in \emph{Proceedings of the 2016 ACM/SIGDA International Symposium on Field-Programmable Gate Arrays}, ser. FPGA ’16.\hskip 1em plus 0.5em minus 0.4em\relax New York, NY, USA: Association for Computing Machinery, Feb 2016, p. 16–25. [Online]. Available: \url{https://doi.org/10.1145/2847263.2847276}
\BIBentrySTDinterwordspacing

\bibitem{9908073}
A.~Nimbekar, C.~S. Vatti, Y.~V.~S. Dinesh, S.~Singh, T.~Gupta, R.~R. Chandrapu, and A.~Acharyya, ``{Low Complexity Reconfigurable-Scalable Architecture Design Methodology for Deep Neural Network Inference Accelerator},'' in \emph{2022 IEEE 35th International System-on-Chip Conference (SOCC)}, 2022, pp. 1--6.

\bibitem{8954866}
J.~Han, Z.~Li, W.~Zheng, and Y.~Zhang, ``{Hardware implementation of spiking neural networks on FPGA},'' \emph{Tsinghua Science and Technology}, vol.~25, no.~4, pp. 479--486, 2020.

\bibitem{Sinharay_2010}
\BIBentryALTinterwordspacing
S.~Sinharay, \emph{\BIBforeignlanguage{en}{Discrete Probability Distributions}}.\hskip 1em plus 0.5em minus 0.4em\relax Elsevier, 2010, p. 132–134. [Online]. Available: \url{https://linkinghub.elsevier.com/retrieve/pii/B9780080448947017218}
\BIBentrySTDinterwordspacing

\bibitem{saluja1987linear}
K.~K. Saluja, ``Linear feedback shift registers theory and applications,'' \emph{Department of Electrical and Computer Engineering, University of Wisconsin-Madison}, pp. 4--14, 1987.

\bibitem{AnalogDevices}
\BIBentryALTinterwordspacing
Pseudo random number generation using linear feedback shift registers. Accessed on 20/10/2023. [Online]. Available: \url{https://www.analog.com/en/design-notes/random-number-generation-using-lfsr.html}
\BIBentrySTDinterwordspacing

\bibitem{he2015deep}
K.~He, X.~Zhang, S.~Ren, and J.~Sun, ``Deep residual learning for image recognition,'' in \emph{Proceedings of the IEEE conference on computer vision and pattern recognition}, 2016, pp. 770--778.

\bibitem{Alizadeh2018AnES}
\BIBentryALTinterwordspacing
M.~Alizadeh, J.~Fern{\'a}ndez-Marqu{\'e}s, N.~D. Lane, and Y.~Gal, ``An empirical study of binary neural networks' optimisation,'' in \emph{International Conference on Learning Representations}, 2018. [Online]. Available: \url{https://api.semanticscholar.org/CorpusID:108364915}
\BIBentrySTDinterwordspacing

\bibitem{le2022adaste}
H.~Le, R.~K. H{\o}ier, C.-T. Lin, and C.~Zach, ``{AdaSTE: An adaptive straight-through estimator to train binary neural networks},'' in \emph{Proceedings of the IEEE/CVF Conference on Computer Vision and Pattern Recognition}, 2022, pp. 460--469.

\bibitem{9756596}
L.~V. Jospin, H.~Laga, F.~Boussaid, W.~Buntine, and M.~Bennamoun, ``Hands-on bayesian neural networks—a tutorial for deep learning users,'' \emph{IEEE Computational Intelligence Magazine}, vol.~17, no.~2, pp. 29--48, 2022.

\bibitem{li2023unleashing}
C.~Li, E.~Jones, and S.~Furber, ``Unleashing the potential of spiking neural networks by dynamic confidence,'' \emph{arXiv preprint arXiv:2303.10276}, 2023.

\bibitem{Jia2020EfficientCR}
\BIBentryALTinterwordspacing
X.~Jia, J.~Yang, R.~Liu, X.~Wang, S.~D. Cotofana, and W.~Zhao, ``Efficient computation reduction in bayesian neural networks through feature decomposition and memorization,'' \emph{IEEE Transactions on Neural Networks and Learning Systems}, vol.~32, pp. 1703--1712, 2020. [Online]. Available: \url{https://api.semanticscholar.org/CorpusID:218562579}
\BIBentrySTDinterwordspacing

\end{thebibliography}
\end{document}